\documentclass[10pt,twocolumn,letterpaper]{article}

\usepackage[pagenumbers]{cvpr} 

\usepackage{setspace}
\usepackage[linesnumbered,ruled,vlined]{algorithm2e}
\usepackage{multirow}
\usepackage{paralist}
\usepackage{multirow}
\usepackage{booktabs}
\usepackage{pifont}
\usepackage{subcaption}
\usepackage{makecell}

\definecolor{cvprblue}{rgb}{0.21,0.49,0.74}

\newcommand{\rf}[1]{{\color{red}{#1}}}
\newcommand{\bd}[1]{{\color{blue}{#1}}}
\usepackage[pagebackref,breaklinks,colorlinks,allcolors=cvprblue]{hyperref}

\usepackage{hyperref}
\usepackage{url}
\usepackage{footnote}

\title{FoundIR: Unleashing Million-scale Training Data to Advance Foundation Models for Image Restoration}

\author{Hao Li$^{*}$ \quad Xiang Chen$^{*}$ \quad Jiangxin Dong \quad Jinhui Tang \quad Jinshan Pan$^{\dagger}$\\
School of Computer Science and Engineering, Nanjing University of Science and Technology\\
{\tt \url{https://www.foundir.net}}
}

\begin{document}


\twocolumn[{%
\renewcommand\twocolumn[1][]{#1}%
\maketitle
\vspace{-10mm}
\begin{center}
     \begin{tabular}{cc}
     \hspace{-3mm}
     \includegraphics[width=0.59\linewidth]{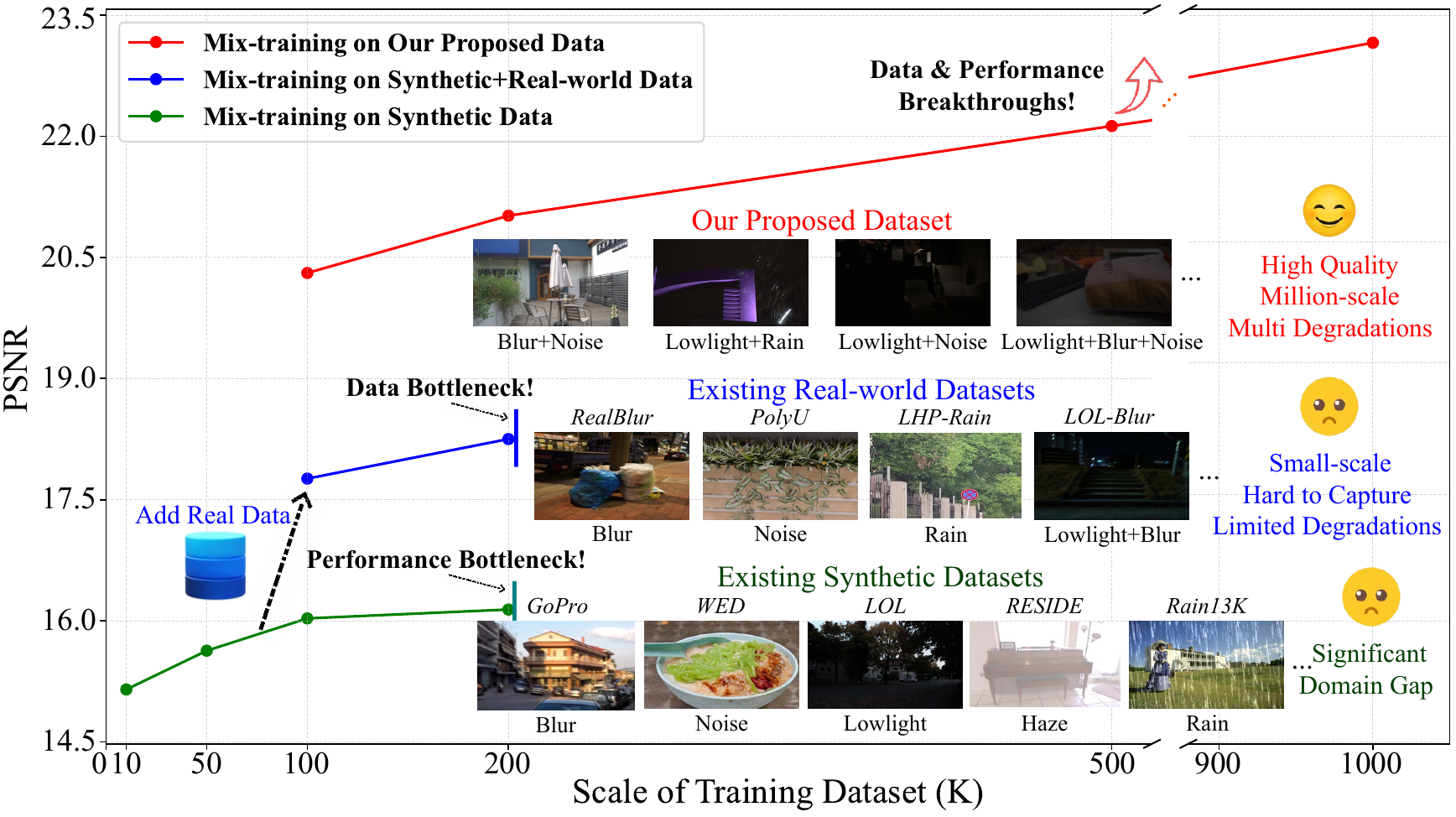} &
     \hspace{-3mm}
     \includegraphics[width=0.4\linewidth]{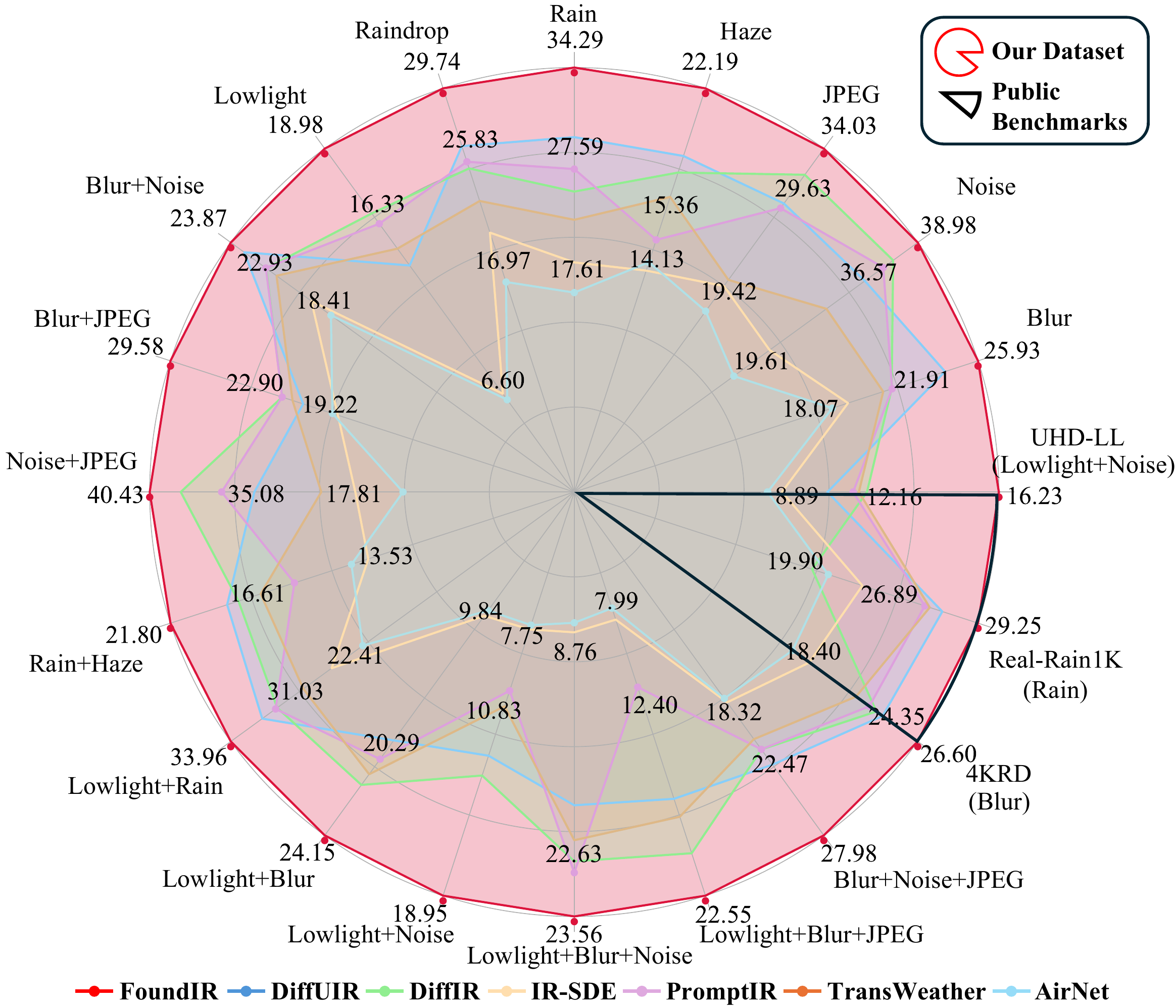} \\
     \hspace{-3mm}
     (a) Scaling up real-world training \textbf{data} &
     \hspace{-3mm}
     (b) Scaling up \textbf{model} generalization
\end{tabular}
     \vspace{-3mm}
     \captionof{figure}{
         The potential of large-scale training data for universal image restoration.
         (a) Analysis of universal image restoration performance in real-world scenarios as training data vary. As the size of real-world training data increases, the image restoration model can achieve significant performance improvement.
         (b) Our proposed FoundIR, trained on our million-scale dataset, achieves state-of-the-art performance across a broad range of image restoration tasks compared to existing universal image restoration methods.
     }
     \vspace{-1mm}
     \label{fig:potential}
 \end{center}%
}]

\def\thefootnote{$*$}\footnotetext{Co-first authorship}
\def\thefootnote{$\dagger$}\footnotetext{Corresponding author}

\begin{abstract}
Despite the significant progress made by all-in-one models in universal image restoration, existing methods suffer from a generalization bottleneck in real-world scenarios, as they are mostly trained on small-scale synthetic datasets with limited degradations.
Therefore, large-scale high-quality real-world training data is urgently needed to facilitate the emergence of foundational models for image restoration.
To advance this field, we spare no effort in contributing a million-scale dataset with two notable advantages over existing training data: real-world samples with larger-scale, and degradation types with higher diversity.
By adjusting internal camera settings and external imaging conditions, we can capture aligned image pairs using our well-designed data acquisition system over multiple rounds and our data alignment criterion.
Moreover, we propose a robust model, FoundIR, to better address a broader range of restoration tasks in real-world scenarios, taking a further step toward foundation models.
Specifically, we first utilize a diffusion-based generalist model to remove degradations by learning the degradation-agnostic common representations from diverse inputs, where incremental learning strategy is adopted to better guide model training.
To refine the model's restoration capability in complex scenarios, we introduce degradation-aware specialist models for achieving final high-quality results.
Extensive experiments show the value of our dataset and the effectiveness of our method.
\end{abstract}

\vspace{-5mm}

\section{Introduction}
\label{sec:intro}
The foundation model~\cite{bommasani2021opportunities} is an emerging paradigm that denotes a model trained on broad data that is capable of being adapted to a wide range of tasks.
Recently, NLP and high-level CV foundation models, such as GPT-4~\cite{achiam2023gpt}, CLIP~\cite{CLIP}, SAM~\cite{SAM}, have achieved remarkable generalization capabilities across a broader scope of scenes.
However, in contrast to the advancements in these fields enabled by extensive datasets and large-scale models, advancements in foundational models for image restoration are less pronounced.

To advance this field, general image restoration methods~\cite{Restormer,SwinIR,lin2024improving} attempt to design a base framework to handle multiple subtasks.
Although these methods can offer better generality compared to task-specific image restoration approaches~\cite{wang2024zero,FFTformer,DRSformer}, they require separate training for different restoration tasks, which is resource-intensive.

Recently, numerous universal (also known as all-in-one) image restoration frameworks~\cite{Airnet,PromptIR,DiffUIR} have been proposed, viewed as potential solutions for becoming foundation models, which aims to simultaneously handle multiple restoration tasks on a single model.
However, these approaches~\cite{Airnet,PromptIR,Prores,DiffIR,DiffUIR} simply combine several public synthetic datasets~\cite{REDS,GoPro,BSD,SIDD,LOL,LOL-Blur,Rain100,RESIDE,LSDIR} as their corresponding all-in-one training sets.
We note that as the scale of synthetic data increases, these methods exhibit significant performance bottlenecks in real-world scenarios (see Figure~\ref{fig:potential}(a)), due to the substantial domain gap between synthetic data and real degradations.
By continuously adding existing real-world datasets~\cite{PolyU,RainDS,LOL-Blur} to the training set, restoration performance can be significantly improved in real-world scenarios.
Unfortunately, the limited amount of existing real-world datasets once again restricts the upper bound of performance, becoming a stumbling block to the development of robust foundational models for image restoration.
Therefore, it is urgent to build a large-scale, high-quality dataset to facilitate the advancement of real-world universal image restoration.

To this end, we make considerable efforts to contribute a million-scale high-quality dataset for image restoration foundational models, consisting of over one million paired high-resolution LQ and HQ images.
For the convenience of paired data acquisition, we build a mechatronic shooting system.
By running this system in multiple rounds, we adjust internal camera settings and external imaging conditions to capture various degradations, and further propose a data alignment strategy to avoid mechanical errors.
Compared to existing training data, our proposed dataset offers twofold advantages: (i) real-world scenarios with larger-scale, and (ii) degradation types with higher-diversity.

Armed with the proposed large-scale dataset, our focus shifts to exploring viable solutions for foundational models in image restoration.
This leads us to two critical questions: (i) how can we formulate a powerful model to handle more real-world degradation types, and based on this, (ii) how can we better train this model when faced with million-scale data?
For the first question, existing methods typically introduce various priors or prompts to guide the model in learning different degradation types.
Unfortunately, these methods place an excessive burden on the model to learn intricate degradation information, leading to increased learning burden and intensified competition among different tasks~\cite{GRIDS}.
For the second question, existing methods typically divide training data into multiple subsets and construct training batches from these subsets for model training.
However, as the scale of training data from different distributions increases, these methods are prone to the common issue of catastrophic forgetting~\cite{zhou2021image,kong2024towards} in machine learning, potentially weakening the model’s performance.

To address the aforementioned issues, we propose a robust model, FoundIR, aiming to make a solid step towards foundation models on image restoration.
Specifically, we first formulate a diffusion-based generalist model to learn degradation-agnostic representations so that the learned features are robust to complex real-world scenarios.
Note that an incremental learning strategy is introduced to overcome potential catastrophic forgetting problem during large-scale data training.
Considering the characteristics of different degradations in real-world scenarios, we further incorporate degradation-aware specialist models to refine the model’s restoration capability for high-quality outputs.
Experimental results on 24 benchmarks show that our study has broken through the ceiling of image restoration (see Figure~\ref{fig:potential}(b)).

In this paper, we summary our contributions as follows:
\begin{compactitem}
\item We collect a million-scale high-quality paired dataset, paving the way for the rise and growth of foundational models on image restoration.
	
\item We introduce FoundIR, a robust high-capacity restoration network that adapts to various degradations, and generalizes well across diverse real-world scenarios.
	
\item We demonstrate the effectiveness of the proposed dataset, and show that our FoundIR achieves favorable performance against state-of-the-art ones.
\end{compactitem}

\section{Related Work}
\label{sec:related}

\begin{figure*}[t]
\centering
\includegraphics[width=0.98\linewidth]{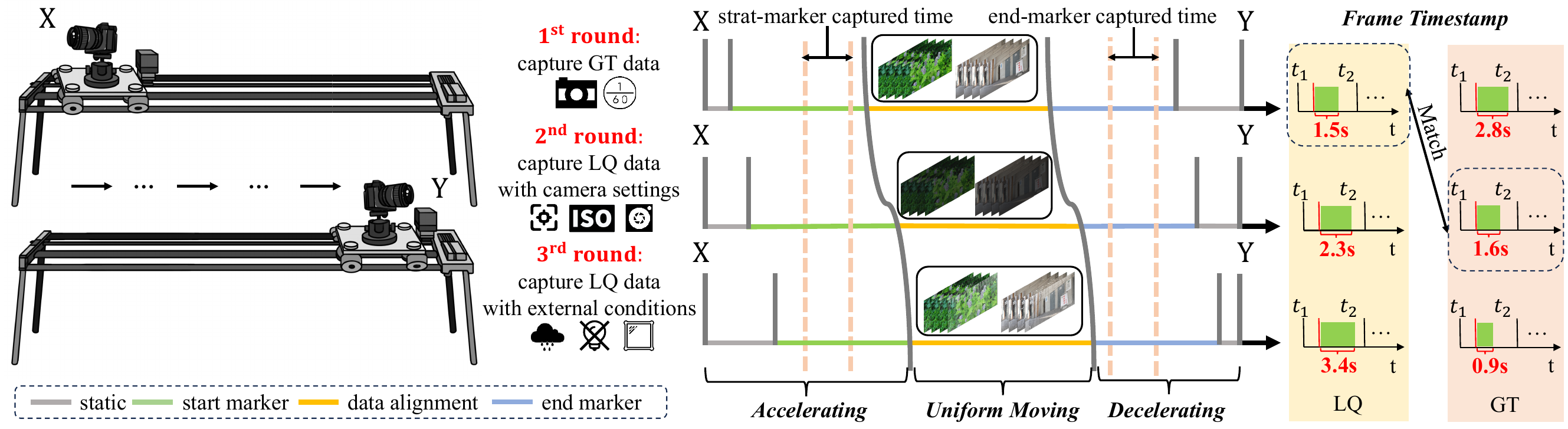}

\vspace{-2mm}
\caption{Illustration of our mechatronic shooting system used for capturing paired data. We move the camera on the electric slide rail from starting point $\mathbf{X}$ to ending point $\mathbf{Y}$.
In the first round, we capture the GT data with the camera set to a fixed exposure time. Then, we capture the LQ data in the second round by adjusting the camera settings (\eg, ISO, aperture, and focus mode), and we capture the LQ data in the third round by changing the external imaging environments (\eg, rain or blocking the light source).
The movement of the camera consists of (\uppercase\expandafter{\romannumeral1}) static phase, (\uppercase\expandafter{\romannumeral2}) accelerating phase, (\uppercase\expandafter{\romannumeral3}) uniform moving phase, and (\uppercase\expandafter{\romannumeral4}) deceleration phase.
Notably, we set up reference objects (\ie, marker) to assist with data alignment to obtain paired images from the uniform moving phase.
}
\label{fig:system}
\vspace{-4mm}
\end{figure*}

{\flushleft\textbf{Universal image restoration}.}
Universal (also known as all-in-one) image restoration aims to handle multiple degradation types simultaneously within a single model~\cite{DiffUIR}.
Compared to task-specific image restoration~\cite{wang2024zero,FFTformer,DRSformer}, universal image restoration poses greater challenges due to the distinct and even mutually exclusive nature of different image degradation factors~\cite{GRIDS,kong2024towards,Onerestore}.
To overcome this challenge, AirNet~\cite{Airnet} introduces contrastive learning to capture discriminative degradation representations.
PromptIR~\cite{PromptIR} and ProRes~\cite{Prores} are proposed to leverage the learnable prompts from the input to guide their networks.
Diffusion-based methods~\cite{DiffIR,DiffUIR,RDDM} are developed to learn the data distributions from various degradations.
Recently, researchers integrate pre-trained large-scale vision models~\cite{DA-CLIP,SUPIR} and multimodal large language models~\cite{MPerceiver,RestoreAgent,SUPIR} into the backbone to facilitate high-fidelity universal image restoration.
However, these approaches depend mostly on the network to learn degradation-relevant information from diverse degraded inputs, leading to increased learning burdens and limited generalization capacity.
In this paper, we drive the model to learn degradation-agnostic common representations for better generalization across real-world scenarios.

\vspace{-2mm}

{\flushleft\textbf{Datasets for universal image restoration}.}
The size and quality of datasets are widely recognized as crucial for universal image restoration.
Existing approaches~\cite{Airnet,PromptIR,Prores,DiffIR,DiffUIR} generally combine several public image restoration datasets~\cite{REDS,GoPro,BSD,SIDD,LOL,LOL-Blur,Rain100,RESIDE} as their training sets, and then evaluate the model on test sets corresponding to specific degradation.
For example, AirNet~\cite{Airnet} and PromptIR~\cite{PromptIR} use a combined training set with different degradations for three image restoration tasks (\eg, BSD400~\cite{BSD} and Urban100~\cite{Urban} for denoising, Rain100L~\cite{Rain100} for deraining, and RESIDE~\cite{RESIDE} for dehazing).
Although training on these combined datasets has advanced the development of universal image restoration methods, relying solely on these limited data (\eg, low quantity and low quality) is far from sufficient to train a robust foundational model for image restoration~\cite{DreamClear}.
We are the first to contribute a million-scale high-quality paired dataset to facilitate the foundation model training and improve the model generalization.

\vspace{-2mm}

{\flushleft\textbf{Incremental learning}.}
Incremental learning, as a basic machine learning paradigm, has been widely applied to various computer vision tasks~\cite{perez2020incremental,cermelli2020modeling,ristin2015incremental} to address the issue of catastrophic forgetting caused by different data distributions during large-scale dataset training.
Existing universal image restoration methods often construct training batches by combining all data or selecting mini-batch from different types of degradation~\cite{Airnet,PromptIR,Prores,DiffIR,DiffUIR,RDDM} for model training.
However, due to variations in input data distributions, training universal image restoration models with these strategies often lead to performance drop (\ie, catastrophic forgetting) as the number of degradation types in large-scale data increases.
Instead, we propose to apply an incremental learning strategy to better train a universal image restoration model driven by large-scale data.

\begin{figure*}[t]
    \begin{center}
    \small
    \begin{tabular}{ccc}
    \includegraphics[width=0.27\linewidth]{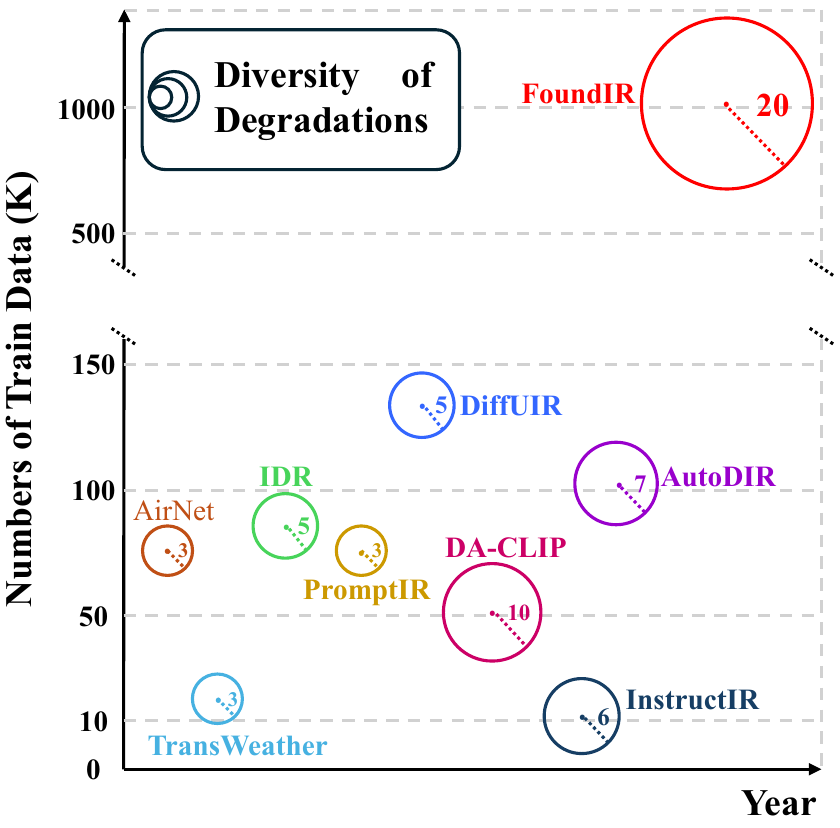} & \includegraphics[width=0.3\linewidth]{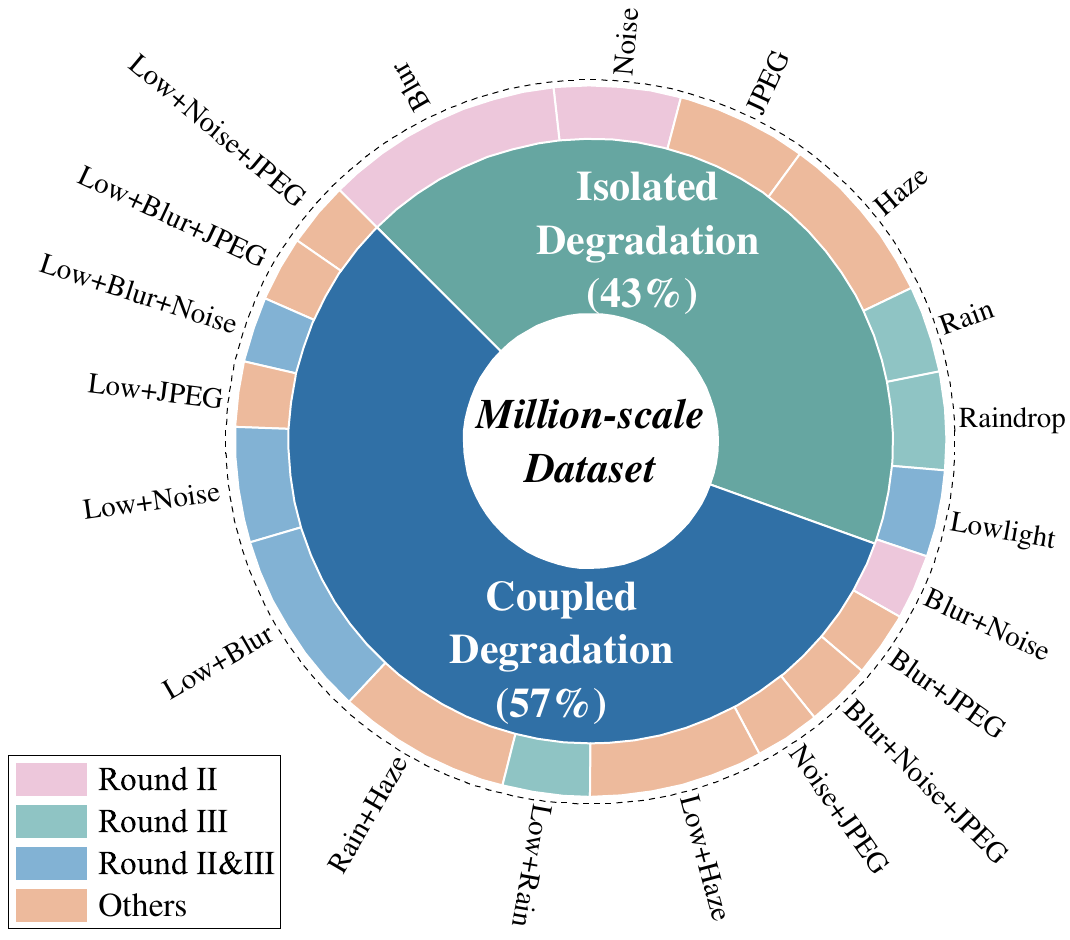} & \includegraphics[width=0.39\linewidth]{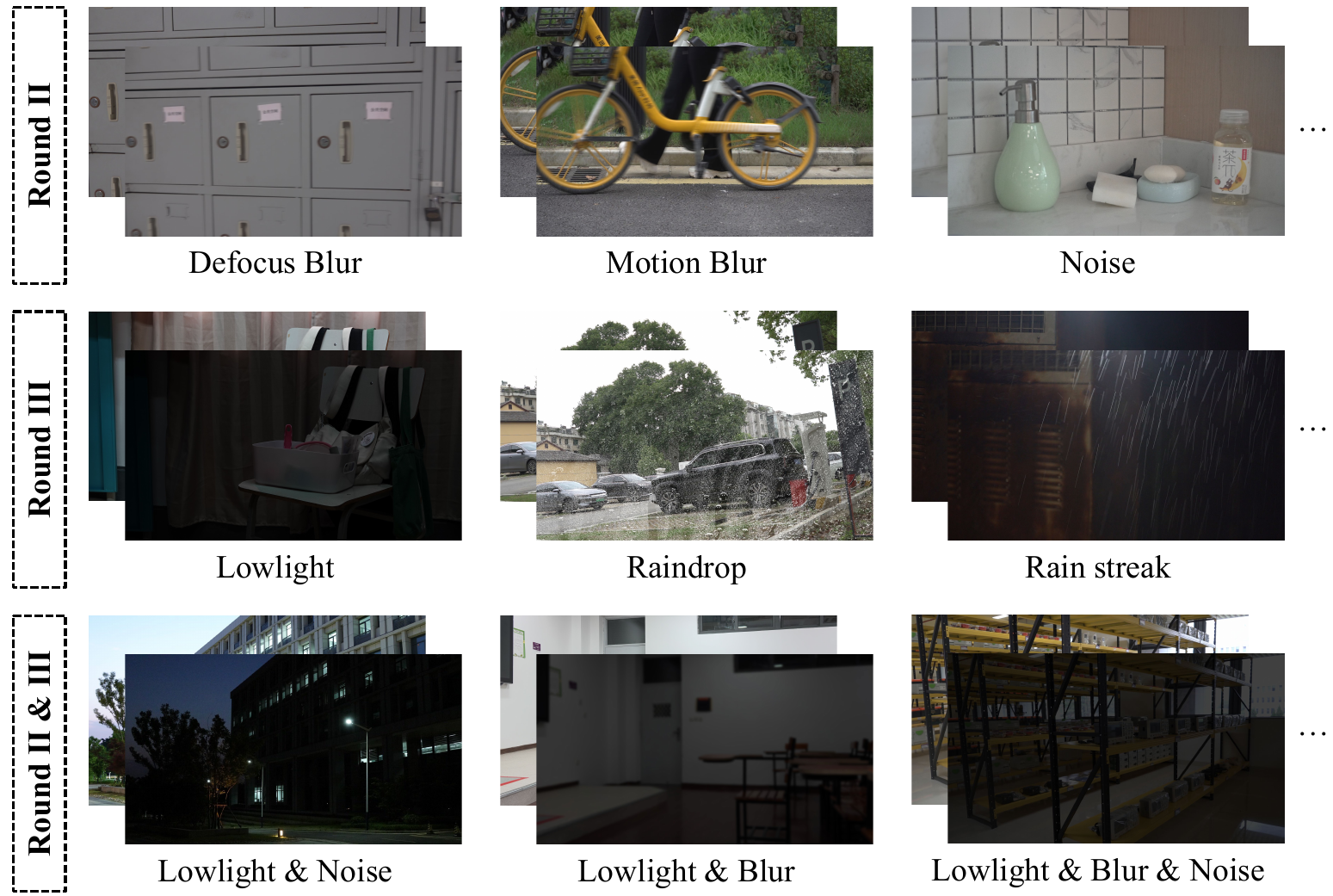} \\
    (a) Scale & (b) Distribution & (c) Diversity
    \end{tabular}
    \vspace{-2mm}
    \captionof{figure}{
        Illustration of the proposed million-scale dataset. (a) Our dataset outperforms existing universal image restoration datasets in terms of training data scale (see y-axis) and the diversity of degradation types (indicated by numbered circles). (b) Distribution of image degradation types of the proposed dataset, including 7 isolated degradation types and 13 coupled degradation types. (c) Example images from our dataset. We adjust internal camera settings (Round \uppercase\expandafter{\romannumeral2}) and external imaging conditions (Round \uppercase\expandafter{\romannumeral3}) to capture various degradation.
    }
    \label{fig:teaser}
\end{center}%
\vspace{-8mm}
\end{figure*}

\section{Million-scale Dataset for Foundation Model}
\label{sec:dataset}
\vspace{-2mm}
To capture large-scale high-quality paired LQ-GT images, a new data collection and an alignment pipeline are proposed and will be detailed in the following sections.

\subsection{Data collection pipeline}
To collect large-scale real-world paired data, we elaborately design a mechatronic shooting system (see Figure~\ref{fig:system}), consisting of a SONY ILCE-7M3 camera, an electric slide rail (GVM Slider 120), and two same tripods.
During the overall shooting process, we use two remote apps (GVM Slider and Imaging Edge Mobile) to control the electric slide rail and camera shutter, respectively.
Here, our data collection pipeline can be divided into three steps: (1) capturing GT data; (2) capturing LQ data with internal camera settings; (3) capturing LQ data with external imaging conditions.

\vspace{-2mm}

{\flushleft\textbf{GT capture.}}
To capture GT data from more diverse scene, we move the camera on the electric slide rail from starting point $\mathbf{X}$ to ending point $\mathbf{Y}$ using the $120^{\circ}$ wide shot mode.
The physical motion trajectory of the camera includes a static phase, an acceleration phase, a uniform motion phase, and a deceleration phase.
Since the camera is unstable in the acceleration and deceleration phases, we only select frames from the static and uniform motion phases as GT images.
In the static phase, to obtain noise-free GT images, we adjust the shutter speed based on ambient illumination and capture GT images directly with a low ISO value ($\mathrm{ISO}\le300$).
In the uniform moving phase, we capture GT sequences using a preset shutter speed (\eg, $\frac{1}{60}$), and carefully select blur-free GT frames in the post-processing of the data.

\vspace{-2mm}

{\flushleft\textbf{LQ capture with internal settings.}}
In real-world scenarios, various degradations (\eg, noise, blur, and low-light) are highly sensitive to camera settings (\eg, ISO Value, and shutter speed (ST)).
To comprehensively capture these LQ images in the second round, we design a series of shooting schemes.
For \textbf{noisy} images, we adjust 30 different camera parameters for each scene (with ISO: $800\sim 20,000$ and ST: $\frac{1}{40}\sim \frac{1}{1000}$).
To ensure consistent illumination across different noisy images of the same scene, we preset a fixed value for the product of ISO value and shutter speed, and all camera settings must strictly adhere to this preset value (\ie, $\mathrm{ISO_1 \times ST_1 \equiv ISO_2 \times ST_2}$).
For \textbf{blurry images caused by defocusing}, we capture the same scene in multiple times by adjusting different focus modes (\eg, AF-S, AF-A, AF-C and MF).
For \textbf{blurry images caused by camera and object movements}, we use a high-frame-rate (240 fps) camera to capture data for simulation.
Following~\cite{MCBlur}, we apply frame interpolation to the captured videos, and blurry images are generated by averaging continuous frames over a time window.
For \textbf{low-light} images, we use the lowest ISO value (\ie, $\mathrm{ISO} = 100$) and set different STs (with $\frac{1}{100}\sim \frac{1}{1000}$) to capture images with varying illumination intensities.

\vspace{-2mm}

{\flushleft\textbf{LQ capture with external conditions.}}
To cover a broader range of real-world degradation data, we capture the LQ data in the third round by changing the external imaging environments.
In fact, image degradation caused by the variation in external factors (\eg, illumination and weather) is more challenging.
For \textbf{ambient illumination}, we capture low-light images in indoor scenes through a series of activities that obstruct light sources, such as turning off the lights and closing the curtains.
Furthermore, considering that image degradation in real-world low-light conditions is characterized by diversity and coupling, we simultaneously add the setup of Round \uppercase\expandafter{\romannumeral2} to capture more complex cases where \textbf{multiple degrading factors coexist}, such as lowlight$+$noise, lowlight$+$blur, lowlight$+$blur$+$noise.

Besides lighting condition, we also introduce external interfering objects to simulate the effects of \textbf{adverse weather} conditions.
The degradation caused by rain in real-world scenarios typically involves rain streaks in the air and raindrops attached to the camera lens or windshields~\cite{RainDS,qian2018attentive}.
To achieve this, we use electric sprinklers to generate \textbf{rain streaks}, a widely-used technique in the Hollywood film industry for simulating rainfall scenes.
To generate \textbf{raindrop} images, we place a glass pane (the thickness of 3 mm) with water droplets in front of the camera.
By stopping the water spray and removing the glass with raindrops, we capture the corresponding rain-free backgrounds.
To ensure data diversity, \textbf{lowlight+rain} is also collected by taking both lighting and weather conditions into account.

\begin{figure*}[t]
    \centering
    \includegraphics[width=\textwidth]{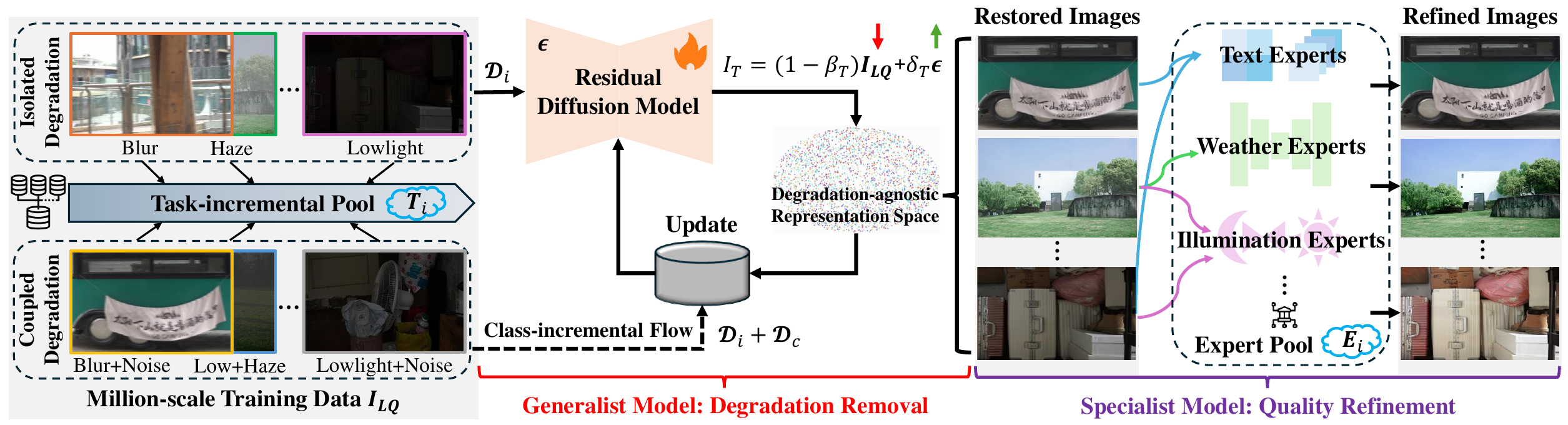}
    \caption{Illustration of the proposed FoundIR. We first employ a diffusion-based generalist model for degradation removal, followed by multiple specialist models for quality refinement. We guide the generalist model to learn a degradation-agnostic common representation space from various degraded inputs, where incremental learning is introduced to improve the model’s training stability. For the specialist models, we construct an expert pool to handle various scenarios, comprising text repair experts, weather experts, and illumination experts.
    }
    \label{fig:network}
    \vspace{-4mm}
\end{figure*}

\vspace{-1mm}

\subsection{Data alignment pipeline}
\vspace{-1mm}
Since we need to run our system for multiple rounds to collect large-scale paired data, ensuring strict spatial alignment of the image sequences is crucial.
The data from the static phase can be naturally utilized as aligned images.
However, the uniform moving phase involves manually selecting the initial and final frames of each sequence, which can easily lead to inevitable bias.
This bias would cause misalignment between each GT-LQ frames, particularly during long-distance movements, where errors can accumulate gradually~\cite{wang2021seeing,fu2023dancing}.
To mitigate this problem, we place recognizable reference objects as \textbf{start-marker} and \textbf{end-marker} before and after the uniform moving phase.
In other words, we manually select aligned GT-LQ frames from the uniform moving phase in a frame-wise manner once the start-marker disappears, continuing until the end-marker appears in the deceleration phase.

Despite the identical slider speed in each round, the captured GT and LQ sequences still exhibit some unpredictable temporal alignment error~\cite{wang2021seeing}.
To address this issue, we introduce a matching alignment strategy, capturing each scene multiple times.
Figure~\ref{fig:system} shows an example of this strategy, we repeatedly capture three GT sequences and three LQ sequences, and calculate the time interval between the appearance and disappearance of the start-marker in each sequence, resulting in six different time intervals.
Then, we match the second GT sequence with the first LQ sequence, which has minimal absolute errors ($0.1s$) of their time intervals.
Using the \textbf{matching alignment} strategy, we ensure that the temporal alignment error of the GT and LQ sequences is below $0.2s$, facilitating accurate alignment.


\subsection{Dataset statistics and features}
Using the above pipelines, we capture around 8,500 scenes in total, including 3,800 indoor scenes and 4,700 outdoor scenes.
We cover 20 types of degradation, including 7 isolated and 13 coupled one.
To visualize the quantity distribution of different degradation types, a sunburst chart is illustrated in Figure~\ref{fig:teaser}(b).
The overall training and testing set of our proposed dataset contains 1,011,614 and 1,500 paired images, respectively.
The average resolution of all images is $2514 \times 1516$.
See Figure~\ref{fig:teaser}(c) for several examples.

The advantages of our benchmark offer two notable advantages over existing universal image restoration datasets.
(i) \textbf{real-world scenarios with larger-scale}: Existing methods combine several small-scale synthetic datasets as training data for universal image restoration (as shown in Figure~\ref{fig:teaser}(a)), limiting real-world applications and foundational model development.
(ii) \textbf{degradation types with higher-diversity}:
Existing training datasets only consider limited isolated degradation types, while in real-world scenarios, image degradation often involves complex interactions and couplings of multiple degradation types.
More details and discussions are provided in the \emph{supplementary materials}.

\section{Proposed Method: FoundIR}
To better deal with the complex and diverse degradations in real-world scenarios, we propose FoundIR, a robust image restoration model trained on our million-scale dataset.
Motivated by ensemble learning in large language models (LLM)~\cite{shao2021intern,lin2024improving,yu2024boosting}, we incorporate one generalist model and multiple specialist models into our FoundIR to form an ensemble framework, which enables better generalization across a broader scope of real-world scenarios while achieving higher restoration quality in specific scenarios.
Figure~\ref{fig:system} presents the overview of the proposed FoundIR method.

\subsection{Degradation-agnostic generalist model}
Firstly, the generalist model is adopted to remove multiple degradations from inputs.
To alleviate the generalist model burden caused by learning specific representations for different degradations, we introduce a degradation-agnostic restoration stage.
This stage aims to formulate a common representation space, where the learned features help remove diverse degradations and improve the robustness of the model in real-world scenarios.
Specifically, we utilize a residual diffusion model to efficiently decouple complex degradation information by transferring the residual components between LQ ($I_{LQ}$) and HQ ($I_{HQ}$) images in the Markov Chain.
Inspired by the residual diffsion process $I_t = I_{t-1}+\alpha_t(I_{LQ}-I_{HQ})+\delta_t \epsilon_{t-1}$ in~\cite{RDDM,DiffUIR}, we introduce additional constraints on the input degraded images $I_{LQ}$, as an explicit condition in the forward process, replacing the traditional representation of $I_t=\epsilon$ in denoising diffusion models (DDPM)~\cite{DDPM}.
Mathematically, the forward diffusion process can calculated as follows:
\vspace{-2mm}
\begin{equation}
\vspace{-2mm}
\begin{aligned}
I_t= & I_{t-1}+\left({\alpha}_t-{\beta}_t\right)I_{LQ}-{\alpha}_tI_{HQ}+\delta_t \epsilon_{t-1}\\
= & I_{t-2} + ({\alpha}_t+{\alpha}_{t-1}-{\beta}_t-{\beta}_{t-1})I_{LQ}\\
&-({\alpha}_t+{\alpha}_{t-1})I_{HQ}+\sqrt{{\delta^2_t}+{\delta^2_{t-1}}}\epsilon_{t-2} \\
= & \cdots \\
= & \left(\bar{\alpha}_t-\bar{\beta}_t\right)I_{LQ}+\left(1-\bar{\alpha}_t\right)I_{HQ}+\bar{\delta}_t \epsilon,\\
\end{aligned}
\label{1}
\end{equation}
where $I_t$ is the output in timestep $t$, and $\epsilon$ denotes gaussian noises. ${\alpha}_t$, ${\beta}_t$, and $\delta_t$ represent the coefficients of the residual term, conditional term, and noise term, respectively.

Notably when $t \rightarrow T$, $\bar{\alpha}_T = 1$, and the Eq.~\eqref{1} could be written as $I_T=\left(1-\bar{\beta}_T\right)I_{LQ}+\bar{\delta}_T \epsilon$.
Here, we assume \(\bar{\gamma}_T = 1- \bar{\beta}_T\) as an adjustable parameter to control the magnitude of degradation-agnostic learning, where \(\bar{\gamma}_t \in [0,1]\).
As noise (\ie, the value of \(\bar{\delta}_t\)) gradually increases during the forward diffusion steps, the features (\ie, the value of \(\bar{\gamma}_t\)) related to the degraded input simultaneously weaken.
By this way, $I_{LQ}$ with different degradations gradually form a common data distribution, facilitating that the model ultimately learns a degradation-agnostic representation space.

\subsection{Incremental learning for model training}
We note that existing universal restoration methods~\cite{Airnet,PromptIR,Prores,DiffUIR} often construct training batches by simply combining all data or selecting mini-batches from different degradation types for training.
However, as the scale of training data increases sharply, the model tends to forget previous knowledge and degrade performance on the early task sequences, making model optimization more difficult~\cite{zhou2021image}.
To alleviate the issue of catastrophic forgetting, we adopt an incremental learning strategy to facilitate better model convergence.
Specifically, we first establish a task-incremental pool $T_{i}$, gradually adding data streams from different tasks to prevent interference and competition~\cite{GRIDS} among multiple tasks during the early model's learning phase.

Unlike merely considering the task-related incremental relationships~\cite{kong2024towards}, we also add a class-incremental flow into the entire training pipeline to better adapt to variable degradations in real-world scenarios.
Here, we divide the training data into two classes: isolated degradation learning class $\mathcal{D}_i$ and coupled degradation learning class $\mathcal{D}_c$.
For class $\mathcal{D}_i$, we first sample batches $\{I^{id}_{LQ},I^{id}_{HQ}\}$ from isolated degradation class for $n$ iterations training, resulting in a model parameter $\theta^{id}$ that learns knowledge of various isolated degradation.
By this way, a well-trained model on isolated degradation class can provide a solid starting point for training on coupled degradation class~\cite{liu2022tape,liu2023degae}.
For class $\mathcal{D}_c$, we then sample combined batches $[\{I^{id}_{LQ},I^{id}_{HQ}\},\{I^{cd}_{LQ},I^{cd}_{HQ}\}]$ from these two classes for $2n$ iterations training, updating the model parameters from $\theta^{id}$ to $\theta$.
The pseudo for training the generalist model is illustrated in Algorithm~\ref{alg:Incremental}.

\setlength{\textfloatsep}{1pt}

\begin{algorithm}[t]
\raggedright
\setstretch{0.95}
	\caption{Incremental Learning.}
	\label{alg:Incremental}
        \KwIn{Isolated degradation class $\mathcal{D}_{i}$; \\
        \quad\quad\quad Coupled degradation class $\mathcal{D}_{c}$; \\
        \quad\quad\quad Noise distribution $\epsilon \sim \mathcal{N}(\mathbf{0,I})$; \\
        \quad\quad\quad Timestep $t \sim Uniform(1,\cdots,T)$; \\
        }
        \KwOut{Model parameter $\theta$;\\
        }

        \For {$T_{i} \in \mathcal{D}_{i}$}{
            Sample batches $\{I^{id}_{LQ},I^{id}_{HQ}\} \in T_i$; \\
            $I^{id}_{HQ}\sim q(I^{id}_{HQ}), R = I^{id}_{LQ}-I^{id}_{HQ}$;\\
            $I_t = \left(\bar{\alpha}_t-\bar{\beta}_t\right)I^{id}_{LQ} +\left(1-\bar{\alpha}_t\right)I^{id}_{HQ}+\bar{\delta}_t \epsilon$; \\
            $\theta^{id} \leftarrow \nabla_{\theta} \Vert R - R_{\theta}(I_t, I^{id}_{HQ}, t) \Vert_1 $;
        }
        \textbf{load} Model parameter $\theta^{id}$;\\
        \For {$T_{i} \in \mathcal{D}_{i}$ and $T_{j} \in \mathcal{D}_c$}{
            Sample batches $[\{I^{id}_{LQ},I^{id}_{HQ}\} \in T_{i},\{I^{od}_{LQ},I^{od}_{HQ}\} \in
 T_{j}]$; \\
            $I^{id/od}_{HQ}\sim q(I^{id/od}_{HQ}), R^{id/od} = I^{id/od}_{LQ}-I^{id/od}_{HQ}$;\\
            $I_t = \left(\bar{\alpha}_t-\bar{\beta}_t\right)I^{id/od}_{LQ} +\left(1-\bar{\alpha}_t\right)I^{id/od}_{HQ}+\bar{\delta}_t \epsilon$; \\
            $\theta \leftarrow \nabla_{\theta} \Vert R^{id} - R^{id}_{\theta}(I_t, I^{id}_{HQ}, t) \Vert_1 +\nabla_{\theta} \Vert R^{od} - R^{od}_{\theta}(I_t, I^{od}_{HQ}, t) \Vert_1 $;\\
        }
        \textbf{return} Final model parameter $\theta$.
\end{algorithm}

\subsection{Degradation-aware specialist model}
Recent studies~\cite{GRIDS,MOME} have demonstrated that a generalist model typically underperforms compared to a specialist model on certain tasks, which can be attributed to task interference.
To this end, we further introduce several specialist models to refine the partial restoration capability in complex scenarios by considering the characteristics of degradation-aware information.
Similar to~\cite{RestoreAgent}, we automatically select the most suitable model from the expert pool $E_{i}$ for each scenario based on specific degradation patterns in the input images, ensuring high-quality output.
For example, we utilize weather experts~\cite{DRSformer} to boost restoration quality under adverse weather conditions, and illumination experts~\cite{Wave-Mamba} to enhance restoration quality under low-light conditions.

Unlike using multiple specialist models for sequential restoration~\cite{RestoreAgent}, our method allows specialist models to share their expertise, collaboratively improving the restoration quality.
Under the guidance of the restoration results by the generalist model, our specialist model can quickly adapt to specific tasks with lower training costs.

\vspace{-1mm}

\section{Experiments}
\vspace{-1mm}
\subsection{Experimental settings}
\vspace{-1mm}
{\flushleft\textbf{Datasets and metrics}.}
We conduct the experiments on our proposed dataset, which contains 20 test sets with different real degradations.
In addition, we evaluate the model's generalization on four public real-world benchmarks, including 4KRD~\cite{4KRD} (deblurring), RealRain-1K~\cite{Real-Rain-1K} (deraining), HazeRD~\cite{HazeRD} (dehazing) and UHD-LL~\cite{UHD-LL} (low-light enhancement and denosing).
We adopt PSNR and SSIM calculated on RGB channel as the evaluation metrics.

\begin{table*}[h!]
\centering
\vspace{-3mm}
\caption{Quantitative comparisons ($\frac{\mathrm{PSNR}}{\mathrm{SSIM}}$) with state-of-the-art general/universal image restoration methods on the proposed benchmark. $\dagger$ denotes evaluation on the benchmark using the corresponding official pretrained model. \rf{Red} and \bd{Blur} indicate the best and the second-best performance. Here, `B', `N', `J', `R', `H', and `L' represent Blur, Noise, JPEG compression, Rain, Haze, and Lowlight, respectively.}
\vspace{-1.5mm}
\resizebox{\textwidth}{!}{%
\begin{tabular}{l|ccccccc||ccccccccccccc||c}
\toprule
\multirow{2}{*}{Methods} & \multicolumn{7}{c||}{Isolated Degradation}  & \multicolumn{13}{c||}{Coupled Degradation} & \multirow{2}{*}{Average} \\  \cline{2-21}
 & Blur & Noise & JPEG & Haze & Rain & Raindrop & Lowlight & B+N & B+J & N+J & R+H & L+H & L+R & L+B & L+N & L+J & L+B+N & L+B+J & L+N+J & B+N+J \\ \hline
 \multirow{2}{*}{Real-ESRGAN~\cite{Real-esrgan}} & 25.20 & 34.46 & 27.62 & \bd{22.07} & 28.95 & \bd{28.94}  & 19.26 & 23.48 & 20.41 & 29.71 & \bd{20.40} & \bd{21.79} & 32.16 & 21.95 & \bd{17.49} & 22.89 & 21.43 & 19.49 & 26.11 & 21.40 & 24.26
 \\
 & 0.7868 & 0.9585 & 0.9108 & \bd{0.8380} & 0.9226 & \bd{0.9115} & \rf{0.8709} & 0.7728 & 0.6562 & 0.9566 & 0.7153 & \rf{0.8084} & 0.9116 & \bd{0.7254} & \bd{0.7358} & 0.8704 & 0.6612 & 0.6582 & 0.9496 & 0.7197 & \bd{0.8173} \\ \hline

 \multirow{2}{*}{AirNet~\cite{Airnet}}  & 18.07 & 19.61 & 19.42 & 14.13 & 17.61 & 16.97 & 6.60 & 18.41 & 19.22 & 17.81 & 13.53 & 8.39 & 22.41 & 9.84 & 7.75 & 8.02 & 8.76 & 7.99 & 7.83 & 18.32 & 14.03 \\
 & 0.5950 & 0.7588 & 0.5495 & 0.3792 & 0.4446 & 0.3233 & 0.1153 & 0.6054 & 0.6108 & 0.5239 & 0.2376 & 0.1452 & 0.6397 & 0.3293 & 0.2162 & 0.2210 & 0.2676 & 0.3122 & 0.3065 & 0.5282 & 0.4054 \\ \hline

 \multirow{2}{*}{DGUNet~\cite{DGUNet}} & 21.86 & 36.69 & 29.80 & 18.58 & 25.47 & 24.83 & 16.16 & 22.90 & 22.86 & 35.28 & 18.79 & 13.20 & 30.66 & 11.54 & 12.56 & 12.17 & 10.60 & 9.57 & 9.64 & 22.43 & 20.27 \\
 & 0.7334 & 0.9494 & 0.8906 & 0.6762 & 0.8245 & 0.7924 & 0.6494 & 0.7633 & 0.7358 & 0.9564 & 0.4981 & 0.4182 & 0.9097 & 0.4922 & 0.5349 & 0.5433 & 0.4608 & 0.4897 & 0.5456 & 0.6945 & 0.6779 \\ \hline

 \multirow{2}{*}{Restormer~\cite{Restormer}} & 21.81 & 34.80 & 29.08 & 13.53 & 24.70 & 23.87 & 8.64 & 22.83 & 22.76 & 34.09 & 12.88 & 10.84 & 29.72 & 10.90 & 8.40 & 10.08 & 9.74 & 9.30 & 10.30 & 22.28 & 18.52 \\
 & 0.7268 & 0.9311 & 0.8742 & 0.5316 & 0.7773 & 0.7395 & 0.3845 & 0.7474 & 0.7266 & 0.9412 & 0.4096 & 0.3110 & 0.8757 & 0.4487 & 0.3255 & 0.4492 & 0.4110 & 0.4647 & 0.5761 & 0.6836 & 0.6167 \\ \hline

  \multirow{2}{*}{TransWeather~\cite{Transweather}} & 21.34 & 30.12 & 23.52 & 17.72 & 23.49 & 22.94 & 14.95 & 22.19 & 22.14 & 25.59 & 18.43 & 15.99 & 28.29 & 21.35 & 11.56 & 19.16 & 20.83 & 19.64 & 21.41 & 21.64 & 21.11 \\
  & 0.7103 & 0.8945 & 0.7296 & 0.5635 & 0.6886 & 0.6814 & 0.6295 & 0.7363 & 0.7105 & 0.8377 & 0.4245 & 0.4285 & 0.8663 & 0.6764 & 0.5181 & 0.6528 & 0.6607 & 0.6918 & 0.8068 & 0.6577 & 0.6782 \\ \hline

 \multirow{2}{*}{IDR~\cite{IDR}} & 17.75 & 21.51 & 17.96 & 12.60 & 16.37 & 16.21 & 6.96 & 18.05 & 18.21 & 16.05 & 11.40 & 9.77 & 23.42 & 10.34 & 7.92 & 8.32 & 9.51 & 8.89 & 8.34 & 17.16 & 13.83 \\
 & 0.6139 & 0.7896 & 0.5207 & 0.3581 & 0.4363 & 0.3422 & 0.1837 & 0.6243 & 0.6063 & 0.4826 & 0.2410 & 0.2377 & 0.7939 & 0.3976 & 0.2631 & 0.2696 & 0.3652 & 0.4170 & 0.3500 & 0.5209 & 0.4406 \\ \hline

 \multirow{2}{*}{PromptIR~\cite{PromptIR}} & 21.91 & 36.57 & 29.63 & 15.36 & 27.59 & 25.83 & 16.33 & 22.93 & 22.90 & 35.08 & 16.61 & 15.36 & 31.03 & 20.29 & 10.83 & 21.52 & \bd{22.63} & 12.40 & 22.71 & 22.47 & 22.49 \\
 & 0.7339 & 0.9478 & 0.8842 & 0.4982 & 0.8411 & 0.8327 & 0.6353 & 0.7587 & 0.7360 & 0.9524 & 0.4985 & 0.4982 & 0.9039 & 0.6723 & 0.4499 & 0.7671 & \bd{0.6924} & 0.5597 & 0.8616 & 0.6942 & 0.7209 \\ \hline

 \multirow{2}{*}{DiffIR~\cite{DiffIR}} & 21.88 & 37.69 & 32.94 & 19.06 & 25.77 & 25.34 & 16.90 & 22.90 & \bd{22.91} & 38.97 & 19.63 & 18.78 & 30.87 & \bd{22.10} & 14.81 & \bd{25.66} & 22.00 & \bd{21.68} & \bd{31.36} & 22.48 & \bd{24.68} \\
 & 0.7346 & 0.9508 & 0.9313 & 0.7801 & 0.8249 & 0.8183 & 0.7311 & 0.7502 & \bd{0.7373} & 0.9732 & 0.6275 & 0.7435 & 0.8971 & 0.7203 & 0.4651 & \bd{0.8761} & 0.6862 & \bd{0.7303} & \bd{0.9613} & 0.6958 & 0.7817 \\ \hline

 \multirow{2}{*}{IR-SDE~\cite{IR-SDE}} & 19.08 & 23.67 & 21.91 & 13.67 & 20.05 & 20.61 & 6.93 & 19.77 & 18.82 & 22.35 & 12.62 & 9.64 & 25.49 & 10.21 & 7.95 & 8.44 & 9.29 & 8.63 & 8.29 & 18.64 & 15.30 \\
 & 0.6395 & 0.8610 & 0.7738 & 0.5080 & 0.6541 & 0.6789 & 0.1578 & 0.6610 & 0.6118 & 0.8187 & 0.3905 & 0.2664 & 0.7651 & 0.3399 & 0.2049 & 0.2649 & 0.3274 & 0.3729 & 0.4027 & 0.5838 & 0.5141 \\ \hline

 \multirow{2}{*}{RDDM~\cite{RDDM}} & 16.37 & 14.94 & 17.95 & 11.12 & 15.55 & 15.47 & 13.16 & 15.82 & 16.75 & 18.23 & 10.52 & 13.87 & 12.53 & 17.26 & 13.39 & 14.09 & 17.34 & 15.98 & 16.88 & 17.76 & 15.24 \\
 & 0.7183 & 0.8519 & 0.8419 & 0.6435 & 0.7118 & 0.6528 & 0.7027 & 0.7156 & 0.6822 & 0.9066 & 0.4568 & 0.3972 & 0.5537 & 0.6471 & 0.6025 & 0.5931 & 0.6457 & 0.6822 & 0.8222 & 0.6679 & 0.6747 \\ \hline

 \multirow{2}{*}{DiffUIR~\cite{DiffUIR}} & \bd{25.31} & 34.48 & 30.09 & 19.97 & 30.17 & 26.98 & 14.02 & \rf{24.44} & 21.36 & 31.96 & 20.24 & 19.31 & \bd{32.35} & 18.97 & 13.88 & 19.98 & 18.89 & 18.64 & 20.66 & \bd{23.72} & 23.27 \\
 & \rf{0.7979} & 0.9622 & \bd{0.9390} & 0.8193 & \bd{0.9350} & 0.8908 & 0.7101 & \rf{0.8039} & 0.6915 & 0.9782 & \bd{0.7268} & 0.7745 & \bd{0.9255} & 0.6939 & 0.6262 & 0.8706 & 0.6705 & 0.6751 & 0.9338 & \rf{0.7861} & 0.8105 \\ \hline

 \multirow{2}{*}{DA-CLIP~\cite{DA-CLIP}} & 20.92 & 29.45 & 25.77 & 15.91 & 23.04 & 20.86 & 17.34 & 22.15 & 21.36 & 26.03 & 13.97 & 12.62 & 21.58 & 16.17 & 15.70 & 15.86 & 15.30 & 12.46 & 15.34 & 21.19 & 19.15 \\
 & 0.6954 & 0.9027 & 0.7816 & 0.5399 & 0.6849 & 0.6490 & 0.7412 & 0.7293 & 0.6940 & 0.8602 & 0.3563 & 0.3220 & 0.7246 & 0.6241 & 0.6365 & 0.6048 & 0.6202 & 0.5774 & 0.7068 & 0.6589 & 0.6554 \\ \hline

 \multirow{2}{*}{X-Restormer~\cite{X-Restormer}} & 21.88 & 35.86 & 28.58 & 16.47 & 25.70 & 26.11  & 16.02 & 22.90 & 22.85 & 33.83 & 15.17 & 14.75 & 30.93 & 19.38 & 9.75 & 18.66 & 22.00 & 13.63 & 16.40 & 22.41 & 21.66 \\
 & 0.7325 & 0.9412 & 0.8646 & 0.6257 & 0.8076 & 0.8368 & 0.6404 & 0.7544 & 0.7321 & 0.9405 & 0.4354 & 0.4416 & 0.9026 & 0.6605 & 0.3866 & 0.7205 & 0.6841 & 0.5738 & 0.7810 & 0.6894 & 0.7075 \\ \hline \hline

 \multirow{2}{*}{SUPIR$\dagger$~\cite{SUPIR}} & 20.92 & 34.11 & 24.52 & \multirow{2}{*}{/} & \multirow{2}{*}{/} & \multirow{2}{*}{/} & \multirow{2}{*}{/} & 21.69 & 21.34 & 26.97 & \multirow{2}{*}{/} & \multirow{2}{*}{/} & \multirow{2}{*}{/} & \multirow{2}{*}{/} & \multirow{2}{*}{/} & \multirow{2}{*}{/} & \multirow{2}{*}{/} & \multirow{2}{*}{/} & \multirow{2}{*}{/} & 20.18 & \multirow{2}{*}{/} \\
 & 0.6649 & 0.9035 & 0.7189 &  &  &  &  & 0.6828 & 0.6293 & 0.8003 &  &   &  &  &  &  &  &  &  & 0.5768 &  \\ \hline

 \multirow{2}{*}{InstructIR$\dagger$~\cite{InstructIR}} & 20.15 & \bd{38.58} & \bd{33.44} & 16.85 & \bd{30.18} & 21.05 & \bd{20.04} & 21.70 & 21.39 & \bd{39.90} & 13.49 & 13.52 & 29.87 & 17.43 & 16.37 & 18.06 & 12.78 & 17.39 & 19.13 & 22.42 & 22.18 \\
 & 0.6801 & \bd{0.9628} & 0.9378 & 0.7555 & 0.8997 & 0.6828 & 0.8542 & 0.7185 & 0.6814 & \bd{0.9770} & 0.5535 & 0.4983 & 0.8866 & 0.6676 & 0.4625 & 0.7787 & 0.5392 & 0.7092 & 0.9105 & 0.6980 & 0.7426 \\ \hline

 \multirow{2}{*}{AutoDIR$\dagger$~\cite{AutoDIR}}  & 20.31 & 36.84 & 32.99 & 15.23 & 25.69 & 20.82 & \rf{21.90} & 21.90 & 22.03 & 37.55 & 14.90 & 14.50 & 27.15 & 19.14 & 17.49 & 16.14 & 18.91 & 16.91 & 18.78 & 22.37 & 22.07 \\
 & 0.6946 & 0.9221 & 0.9280 & 0.6264 & 0.7711 & 0.6687 & 0.8288 & 0.7293 & 0.7088 & 0.9616 & 0.4213 & 0.4456 & 0.8576 & 0.6570 & 0.6900 & 0.6375 & 0.6499 & 0.6761 & 0.8122 & 0.6876 & 0.7187 \\ \hline \hline

 \multirow{2}{*}{FoundIR} & \rf{25.93} & \rf{38.98} & \rf{34.03} & \rf{22.19} & \rf{34.29} & \rf{29.74} & 18.98 & \bd{23.87} & \rf{29.58} & \rf{40.43} & \rf{21.80} & \rf{22.82} & \rf{33.96} & \rf{24.15} & \rf{18.95} & \rf{29.08} & \rf{23.56} & \rf{22.55} & \rf{33.46} & \rf{27.98} & \rf{27.81} \\
 & \bd{0.7914} & \rf{0.9647} & \rf{0.9427} & \rf{0.8492} & \rf{0.9434} & \rf{0.9160} & \bd{0.8576} & \bd{0.7737} & \rf{0.8544} & \rf{0.9795} & \rf{0.7582} & \bd{0.8048} & \rf{0.9330} & \rf{0.7808} & \rf{0.7454} & \rf{0.9098} & \rf{0.7550} & \rf{0.8181} & \rf{0.9636} & \bd{0.7452} & \rf{0.8529} \\

\bottomrule
\end{tabular}
}
\label{tab:comparison}
\end{table*}

\begin{table*}[h!]
\centering
\vspace{-3mm}
\caption{Quantitative comparisons ($\frac{\mathrm{PSNR}}{\mathrm{SSIM}}$) on the public benchmarks. Where $\dagger$ denotes using the official pretrained model for testing, and `/' indicates that the one was not trained on the corresponding task. \rf{Red} and \bd{Blur} indicate the best and the second-best performance.}
\vspace{-1.5mm}
\resizebox{\textwidth}{!}{%
\begin{tabular}{l|c|c|c|c||c|c|cc|cc|cc|cc||c}
\toprule
\multirow{2}{*}{Methods} & \multicolumn{4}{c||}{General Image Restoration Methods} & \multicolumn{10}{c||}{Universal Image Restoration Methods} & \multirow{2}{*}{FoundIR} \\ \cline{2-15}

& DGUNet~\cite{DGUNet} & Restormer~\cite{Restormer} & IDR~\cite{IDR} & IR-SDE~\cite{IR-SDE} & TransWeather~\cite{Transweather} & AutoDIR$\dagger$~\cite{AutoDIR} & AirNet$\dagger$~\cite{Airnet} & AirNet & PromptIR$\dagger$~\cite{PromptIR} & PromptIR & DiffIR$\dagger$~\cite{DiffIR} & DiffIR  & DiffUIR$\dagger$~\cite{DiffUIR} & DiffUIR   &
 \\ \hline
 \multirow{2}{*}{4KRD~\cite{4KRD}} & 24.33 & 24.32 & 16.66 & 19.69 & 23.32 & 24.92 & \multirow{2}{*}{/} & 18.40 & \multirow{2}{*}{/} & 24.35 & 24.06 & 24.90 & 25.35 & \bd{25.38} & \rf{26.60} \\
  & 0.7347 & 0.7300 & 0.5783 & 0.6468 & 0.7043 & 0.7655 &  & 0.5986 &  & 0.7341 & 0.7477 & 0.7651 & \bd{0.7991} & 0.7894 & \rf{0.8105}  \\ \hline

  \multirow{2}{*}{RealRain-1K~\cite{Real-Rain-1K}} & 26.41 & 26.42 & 22.92 & 22.37 & 27.25 & 22.90 & 18.34 & 19.90 & 22.38 & 26.89 & \multirow{2}{*}{/} & 18.73  & 22.10 & \bd{28.16} & \rf{31.34}  \\
 & 0.8881 & 0.8736 & 0.8129 & 0.8040 & 0.8886 & 0.7938 & 0.6346 & 0.6901 & 0.7615 & 0.8949  &  & 0.5919 & 0.7608 & \bd{0.9293} & \rf{0.9683}  \\ \hline

  \multirow{2}{*}{HazeRD~\cite{HazeRD}} & 15.31 & 14.07 & 12.64 & 13.97 & \bd{16.82} & 15.74 & 14.44 & 14.59 & 15.67 & 16.29 & \multirow{2}{*}{/} & 15.62 & 14.28 & 15.42 & \rf{20.73}  \\
 & 0.7776 & 0.7353 & 0.6390 & 0.7047 & 0.7850 & 0.8285 & 0.6577 & 0.6692 & 0.7869 & 0.6912  &  & \bd{0.8412} & 0.8338 & 0.8397 & \rf{0.9107}  \\ \hline

  \multirow{2}{*}{UHD-LL~\cite{UHD-LL}} & 10.51 & 10.22 & 9.38 & 9.44 & 12.38 & \bd{16.10} & \multirow{2}{*}{/} & 8.89 & \multirow{2}{*}{/} & 12.16  & \multirow{2}{*}{/} & 12.69  & 9.17 & 11.18  & \rf{16.23}  \\
 & 0.5518 & 0.5280 & 0.4576 & 0.4217 & 0.6437 & \bd{0.7149} &  & 0.3932 &  & 0.6273  &  & 0.5986  & 0.3899 & 0.5929  & \rf{0.7626}  \\

\bottomrule
\end{tabular}
}
\label{tab:public}
\vspace{-5mm}
\end{table*}

\vspace{-1mm}

{\flushleft\textbf{Implementation details}.}
We utilize the Adam~\cite{kingma2014adam} optimizer for $2 \times 10^6$ iterations training.
We set the batch size to 80, the patch size to $256 \times 256$, and $\bar{\gamma}_T$ to 0.3.
The initial learning rate is set to $1 \times 10^{-4}$, and reduces to $5 \times 10^{-5}$ after $1 \times 10^6$ iterations.
The same diffusing reverse process and loss function~\cite{DiffUIR} are adopted in our model.
For testing, we first crop the entire image into $1024\times1024$ patches as input and then stitch them back after restoration, and we use 4 timesteps for all tasks.
All experiments are conducted on four servers, each equipped with 8 NVIDIA RTX 4090 GPUs, using PyTorch 2.0.
The proposed dataset, training code and test models are available to \href{https://www.foundir.net}{FoundIR}.

\vspace{-1mm}

\subsection{Comparisons with the state of the art}
\vspace{-1mm}
We compare our proposed FoundIR with 16 recent general / universal image restoration methods, including Real-ESRGAN~\cite{Real-esrgan}, AirNet~\cite{Airnet}, DGUNet~\cite{DGUNet}, Restormer~\cite{Restormer}, TransWeather~\cite{Transweather}, IDR~\cite{IDR}, PromptIR~\cite{PromptIR}, DiffIR~\cite{DiffIR}, IR-SDE~\cite{IR-SDE}, RDDM~\cite{RDDM}, DiffUIR~\cite{DiffUIR}, DA-CLIP~\cite{DA-CLIP}, X-Restormer~\cite{X-Restormer}, SUPIR~\cite{SUPIR}, InstructIR~\cite{InstructIR}, and AutoDIR~\cite{AutoDIR}.
Apart from some methods (\ie, SUPIR~\cite{SUPIR}, InstructIR~\cite{InstructIR}, and AutoDIR~\cite{AutoDIR}), which are constrained by training resources and their unique methodologies, other methods are re-trained on our proposed dataset.

\vspace{-2mm}

\begin{figure*}[t]
	\footnotesize
	\begin{center}
		\begin{tabular}{c c c c c c c c}
			\multicolumn{3}{c}{\multirow{5}*[48pt]{
            \hspace{-2.5mm} \includegraphics[width=0.42\linewidth,height=0.245\linewidth]{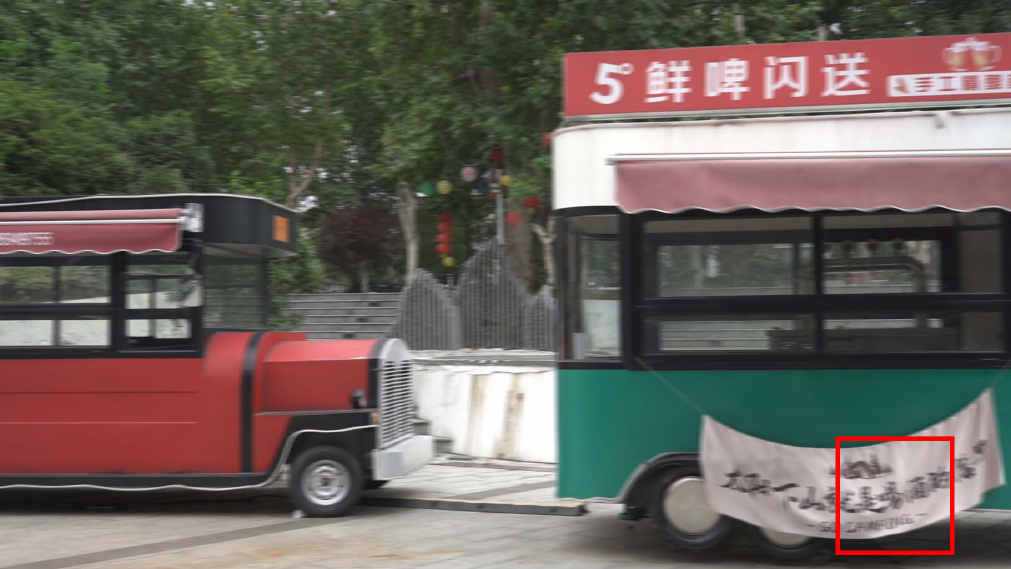}}}
            & \hspace{-4.0mm} \includegraphics[width=0.11\linewidth]{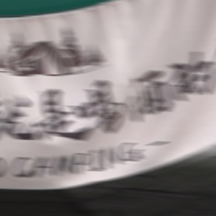}
            & \hspace{-4.0mm} \includegraphics[width=0.11\linewidth]{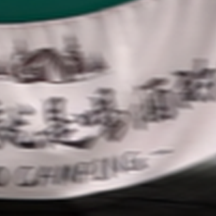}
            & \hspace{-4.0mm} \includegraphics[width=0.11\linewidth]{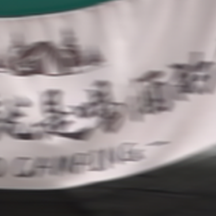}
            & \hspace{-4.0mm} \includegraphics[width=0.11\linewidth]{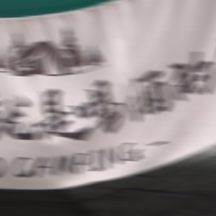}
            & \hspace{-4.0mm} \includegraphics[width=0.11\linewidth]{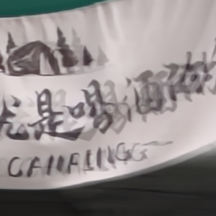}
              \\
		\multicolumn{3}{c}{~}
        & \hspace{-4.0mm} LQ patch
        & \hspace{-4.0mm} AirNet~\cite{Airnet}
        & \hspace{-4.0mm} PromptIR~\cite{PromptIR}
        & \hspace{-4.0mm} DiffIR~\cite{DiffIR}
        & \hspace{-4.0mm} DiffUIR~\cite{DiffUIR} \\		
	\multicolumn{3}{c}{~}
        & \hspace{-4.0mm} \includegraphics[width=0.11\linewidth]{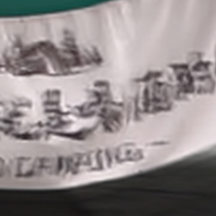}
        & \hspace{-4.0mm} \includegraphics[width=0.11\linewidth]{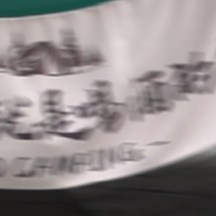}
        & \hspace{-4.0mm} \includegraphics[width=0.11\linewidth]{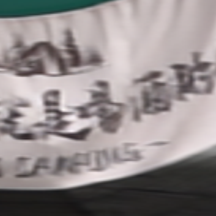}
        & \hspace{-4.0mm} \includegraphics[width=0.11\linewidth]{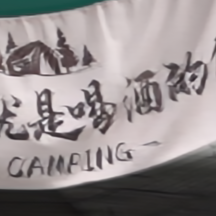}
        & \hspace{-4.0mm} \includegraphics[width=0.11\linewidth]{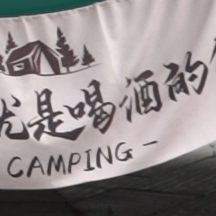}
        \\
	\multicolumn{3}{c}{\hspace{-4.0mm} \textbf{Blur}}
        & \hspace{-4.0mm} DA-CLIP~\cite{DA-CLIP}
        & \hspace{-4.0mm} X-Restormer~\cite{X-Restormer}
        & \hspace{-4.0mm} AutoDIR~\cite{AutoDIR}
        & \hspace{-4.0mm} FoundIR
        & \hspace{-4.0mm} GT patch\\ 	

        \multicolumn{3}{c}{\multirow{5}*[48pt]{
            \hspace{-2.5mm} \includegraphics[width=0.42\linewidth,height=0.245\linewidth]{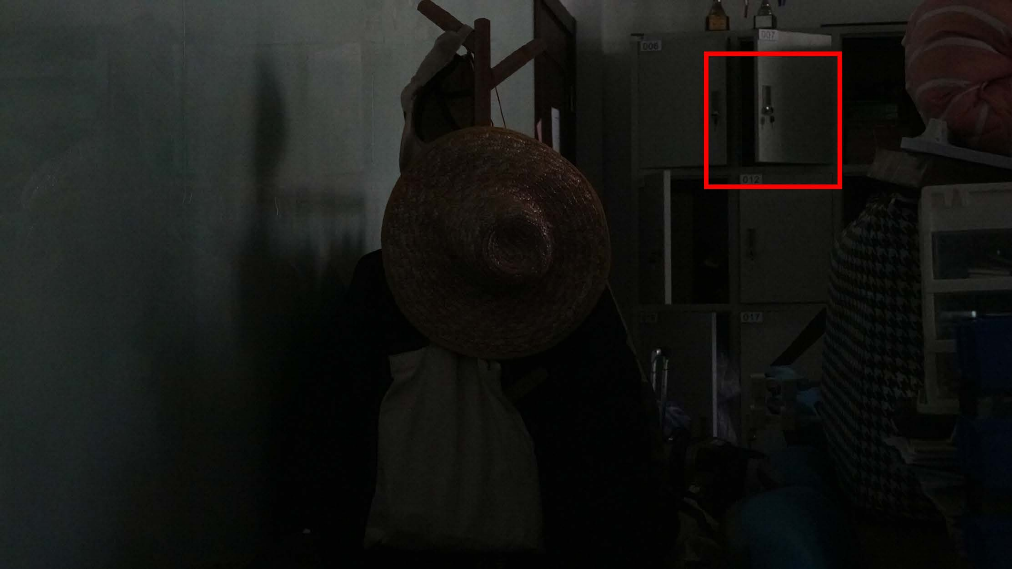}}}
            & \hspace{-4.0mm} \includegraphics[width=0.11\linewidth]{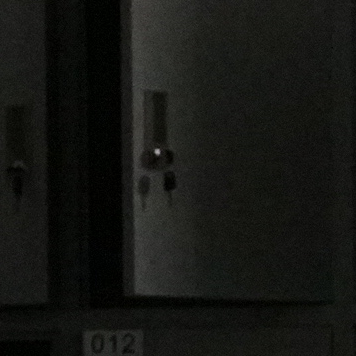}
            & \hspace{-4.0mm} \includegraphics[width=0.11\linewidth]{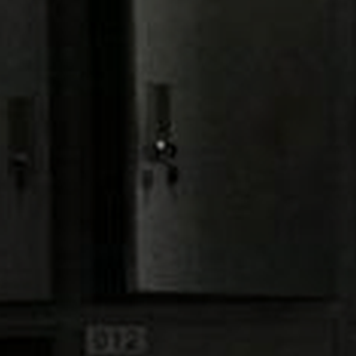}
            & \hspace{-4.0mm} \includegraphics[width=0.11\linewidth]{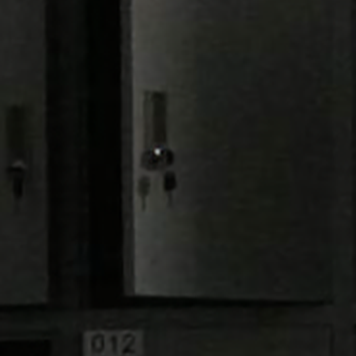}
            & \hspace{-4.0mm} \includegraphics[width=0.11\linewidth]{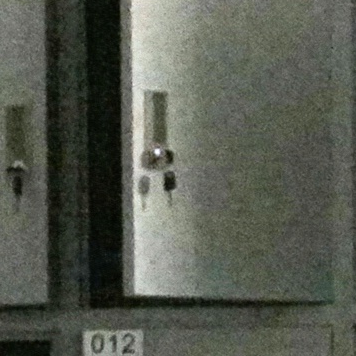}
            & \hspace{-4.0mm} \includegraphics[width=0.11\linewidth]{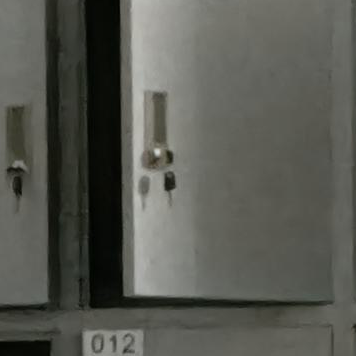}
              \\
		\multicolumn{3}{c}{~}                                   & \hspace{-4.0mm} LQ patch
        & \hspace{-4.0mm} AirNet~\cite{Airnet}
        & \hspace{-4.0mm} PromptIR~\cite{PromptIR}
        & \hspace{-4.0mm} DiffIR~\cite{DiffIR}
        & \hspace{-4.0mm} DiffUIR~\cite{DiffUIR} \\		
	\multicolumn{3}{c}{~}
        & \hspace{-4.0mm} \includegraphics[width=0.11\linewidth]{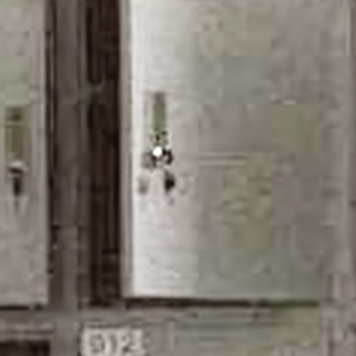}
        & \hspace{-4.0mm} \includegraphics[width=0.11\linewidth]{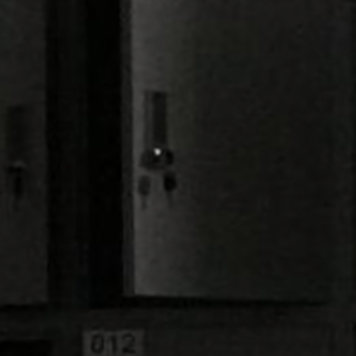}
        & \hspace{-4.0mm} \includegraphics[width=0.11\linewidth]{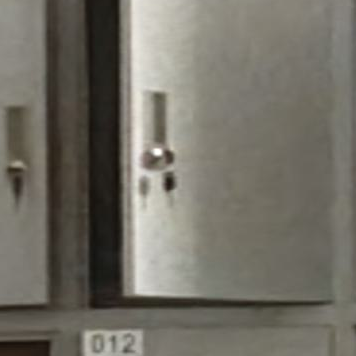}
        & \hspace{-4.0mm} \includegraphics[width=0.11\linewidth]{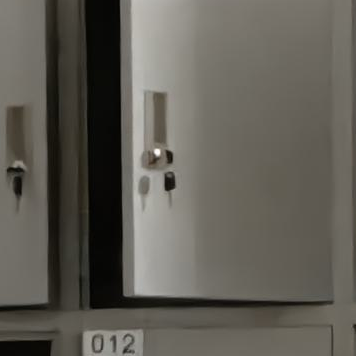}
        & \hspace{-4.0mm} \includegraphics[width=0.11\linewidth]{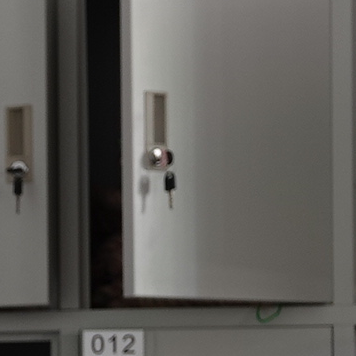}
        \\
	\multicolumn{3}{c}{\hspace{-4.0mm} \textbf{Lowlight+Noise}}
        & \hspace{-4.0mm} DA-CLIP~\cite{DA-CLIP}
        & \hspace{-4.0mm} X-Restormer~\cite{X-Restormer}
        & \hspace{-4.0mm} AutoDIR~\cite{AutoDIR}
        & \hspace{-4.0mm} FoundIR
        & \hspace{-4.0mm} GT patch\\

		\end{tabular}
	\end{center}

	\vspace{-6mm}
	\caption{Visual comparisons on the isolated and coupled degradation inputs from the proposed benchmark. Zoom in for a better view.}
	\label{fig:visual}
	\vspace{-6mm}
\end{figure*}

{\flushleft\textbf{Evaluation on the proposed dataset}.} Table~\ref{tab:comparison} shows the proposed FoundIR achieves the best quantitative results across a broad of image degradation types, including both isolated and coupled degradations.
Compared with the recent method, DA-CLIP~\cite{DA-CLIP}, FoundIR achieves an improvement of 7.98dB in PSNR on the `L+B' category.
Figure~\ref{fig:visual} presents the qualitative comparison results.
Compared to Transformer-based methods (\eg, PromptIR~\cite{PromptIR}) and diffusion-based models (\eg, AutoDIR~\cite{AutoDIR}), our FoundIR restores better structures and details for both isolated and coupled degradation inputs.
These results indicate that our FoundIR has a more comprehensive restoration capability.

\vspace{-2mm}

{\flushleft\textbf{Evaluation on the public benchmarks}.}
We further evaluate the generalization capability of different models on four public benchmarks.
Table~\ref{tab:public} illustrates that our FoundIR still achieves the best restoration performance, indicating its capability to generalize well to out-of-distribution data.
To further demonstrate that training on our proposed dataset allows the model to generalize better to real-world scenarios, we compare the performance of the models trained on our dataset with those trained on existing public datasets.
It is evident that models trained on our dataset consistently improve performance across various tasks (\eg, DiffUIR achieves a 6dB PSNR gain on the RealRain-1K~\cite{Real-Rain-1K} dataset), confirming the generalizability of our dataset.

\vspace{-2mm}

\subsection{Ablation analysis and discussion}

\begin{figure}[t]
    \centering
   \caption{Ablation studies on the data scale and model coefficients.}
   \vspace{-2mm}
  \begin{tabular}{cccc}
  \hspace{-2mm}
  \rotatebox{90}{~~~~~~~~PSNR}

   & \hspace{-6mm}
   \includegraphics[width=0.45\linewidth]{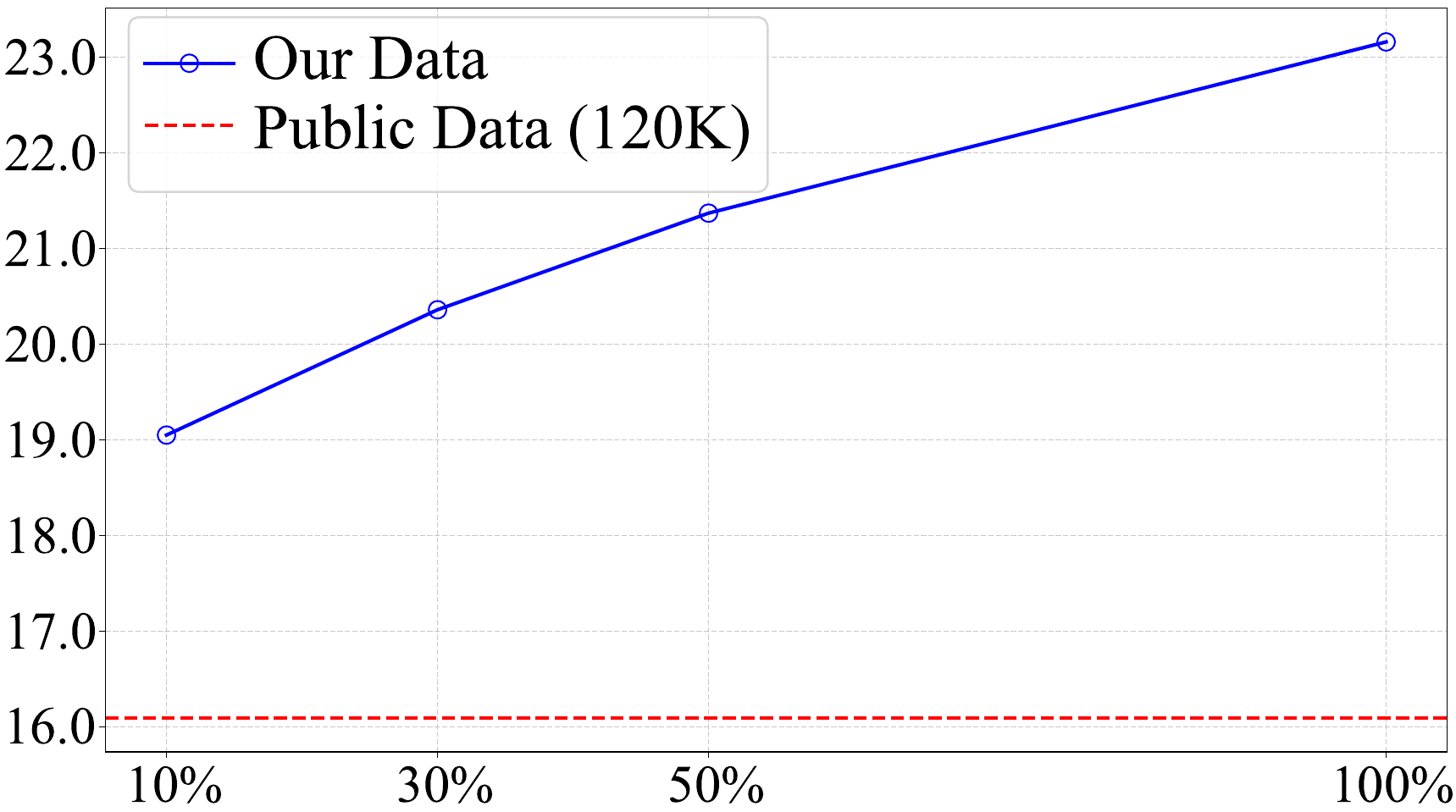}
   & \hspace{-6mm} \rotatebox{90}{~~~~~~~~PSNR}
    &
    \hspace{-6mm}
    \includegraphics[width=0.45\linewidth]{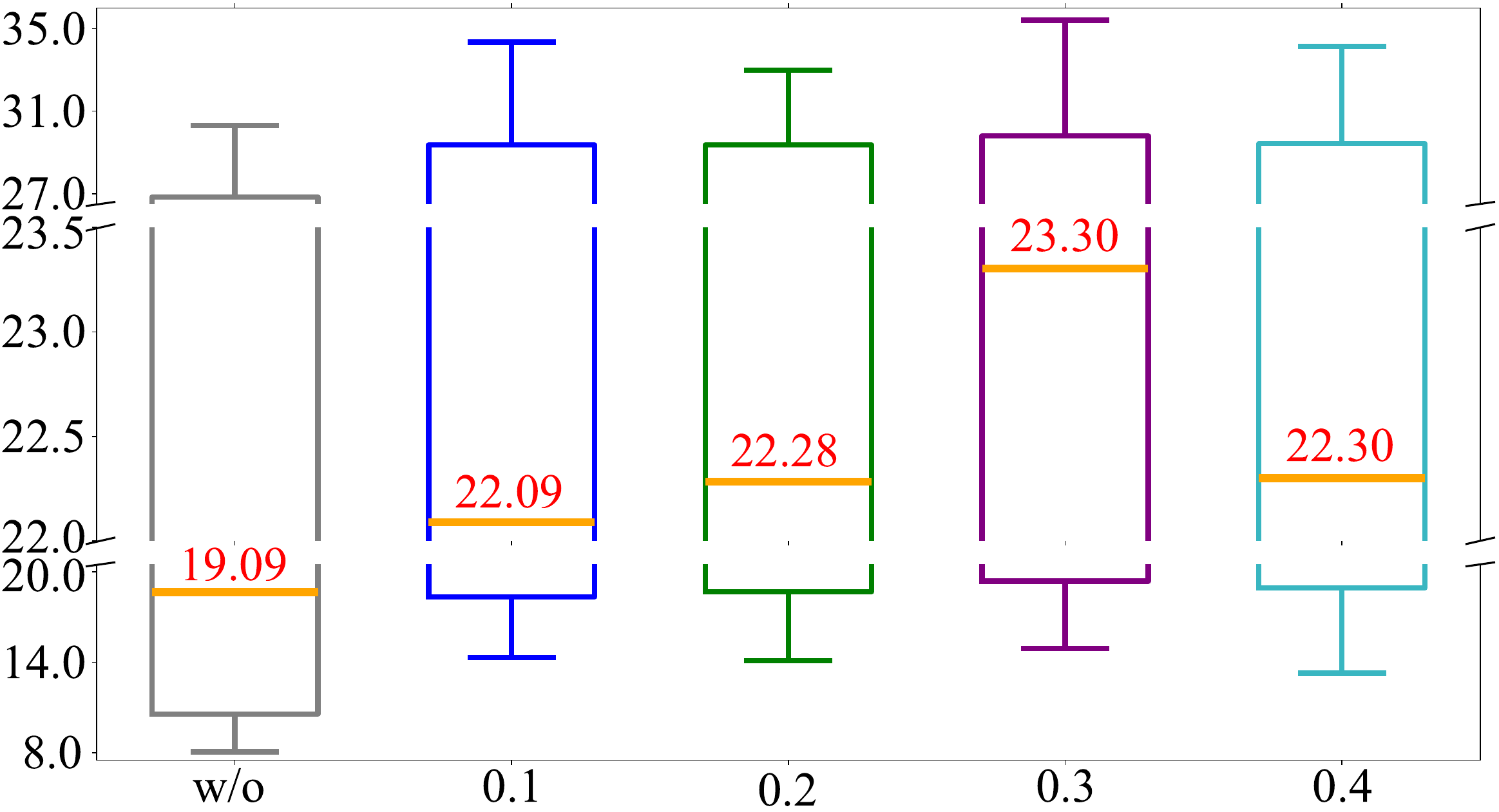}
    \\
   &
    \hspace{-3mm}
    Percentage of training data
    &
    & \hspace{-3mm}
    Condition coefficient $\bar{\gamma}_T$ \\
     \end{tabular}
    \label{fig:ablation}
    \vspace{1mm}
\end{figure}

{\flushleft\textbf{Effect of training data scaling}.}
To investigate the effect of our dataset scale, we gradually increase the percentage of training data from 10\% to 100\% for training.
Figure~\ref{fig:ablation} shows that as the dataset size increases, it achieves higher performance.
We also collect a total of 120,000 LQ-GT image pairs from public real-world datasets~\cite{GoPro,RESIDE,LOL,MCBlur,PolyU} to train our model, using it as the baseline result.
Here, we can find that using 10\% of the proposed dataset achieves a 2.59dB improvement in PSNR compared to baseline, demonstrating that our training data have better generalization than existing datasets under the same scale.

\vspace{-2mm}

{\flushleft\textbf{Effect of degradation-agnostic coefficient}.}
In the residual diffusion model, we set a condition coefficient $\bar{\gamma}_T$ to control the level of degradation-agnostic learning.
We analyze the effect of this coefficient on degradation-agnostic learning.
Figure~\ref{fig:ablation} presents a boxplot of PSNR values for 20 degradation types using different coefficients.
Here, `w/o' means that the model is highly dependent on degradation-relevant features, making it incapable of handling different degradations simultaneously.
Compared to other configurations, `$\bar{\gamma}_T=0.3$' achieves higher PSNR across multiple quantiles (\ie median, maximum, minimum values), indicating that it achieves the best trade-off in handling various degradations.

\vspace{-2mm}

{\flushleft\textbf{Effectiveness of incremental learning}.}
We compare our training strategy with existing approaches used in universal restoration methods, including: (1) directly select training batches from all datasets (Mix-Train)~\cite{PromptIR}, (2) construct training batches by selecting mini-batches from each degradation (Combine-Train)~\cite{DiffUIR}, and (3) task sequential learning (Sequence-Train)~\cite{kong2024towards}.
We also analyze the starting point of our class-incremental flow.
The comparison results in Table~\ref{tab:ablation}(a) show that the model trained with incremental learning ($\mathcal{D}_i \to \mathcal{D}_c$) obtains the best restoration performance.
This provides a feasible solution for large-scale training of foundational models for image restoration.

\begin{table}[t]
    \centering
    \caption{Ablation studies on the training strategy and pipeline.}
    \vspace{-2mm}
    \resizebox{0.49\textwidth}{!}{%
    \begin{tabular}{cc}
     \begin{tabular}{l|cc}
      \toprule
      Strategy & PSNR & SSIM \\ \hline
      Mix-Train & 22.57 & 0.7956 \\
      Combine-Train & 23.40 & 0.8061 \\
      Sequence-Train & 22.71 & 0.7925 \\
      IL ($\mathcal{D}_{c}\to \mathcal{D}_{i}$) & 22.33 & 0.7936 \\
      IL ($\mathcal{D}_{i}\to \mathcal{D}_{c}$) & 23.70 & 0.8201 \\
      \bottomrule
    \end{tabular}
    &
    \begin{tabular}{l|cc}
      \toprule
      Methods & PSNR & SSIM \\ \hline
      Only $\mathcal{G}$ & 18.19 & 0.7505 \\
      Only $\mathcal{S}$ & 13.06 & 0.4729 \\
      $\mathcal{S}_1 \to \mathcal{S}_2$ & 13.52 & 0.4647 \\
      $\mathcal{S} \to \mathcal{G}$ & 15.63 & 0.6312 \\
      $\mathcal{G} \to \mathcal{S}$ & 22.82 & 0.8048 \\
      \bottomrule
    \end{tabular} \\
    \specialrule{0em}{0pt}{4pt}
    (a) Training strategy & (b) Overall pipeline
    \end{tabular}
    }

    \label{tab:ablation}
    \vspace{0.5mm}
\end{table}

\vspace{-3mm}

{\flushleft\textbf{Effectiveness of overall pipeline}.}
To demonstrate the effectiveness of our ensemble framework, we compare the results of different variants, \ie, (1) only $\mathcal{G}$, (2) only $\mathcal{S}$, (3) $\mathcal{S}_1 \to \mathcal{S}_2$~\cite{RestoreAgent}, and (4) $\mathcal{S} \to \mathcal{G}$, on the `L+H' category.
Where $\mathcal{G}$ and $\mathcal{S}$ denote the generalist model and specialist model, respectively.
Table~\ref{tab:ablation}(b) shows that our solution $\mathcal{G} \to \mathcal{S}$ can better improve restoration quality.
More discussions are included in the \textit{supplementary materials}.

\vspace{-2mm}

\section{Conclusion}
\vspace{-1mm}
This paper presents a large-scale high-quality dataset containing one million paired LQ-GT images, serving as valuable training resources for foundation models on universal image restoration.
Furthermore, we propose a robust image restoration model, FoundIR, to address a broader range of real-world degradation scenarios, while leveraging incremental learning techniques to facilitate large-scale data training.
Extensive experiments have demonstrated that the value of our dataset, and the effectiveness of our method.

{
    \small
    \bibliographystyle{ieeenat_fullname}
    \bibliography{main}
}



\end{document}